\documentclass[10pt,twocolumn,letterpaper]{article}

\usepackage{cvpr}              
\usepackage{physics}

\usepackage{floatrow}
\usepackage{multirow}
\usepackage[dvipsnames]{xcolor}

\definecolor{cvprblue}{rgb}{0.21,0.49,0.74}
\usepackage[pagebackref,breaklinks,colorlinks,citecolor=cvprblue]{hyperref}

\def\paperID{19} 
\def\confName{CVPR}
\def\confYear{2024}

\begin{document}
\title{High-fidelity Endoscopic Image Synthesis by Utilizing \\ Depth-guided Neural Surfaces}

\author{Baoru Huang$^{1, 2, *}$, Yida Wang$^{3, *}$, Anh Nguyen$^{4}$, Daniel Elson$^{1}$, Francisco Vasconcelos$^{2}$, Danail Stoyanov$^{2}$ \\
{\small $^{1}$Imperial College London, UK} 
{\small $^{2}$University College London, UK} 
{\small $^{3}$Technical University of Munich, Germany} \\
{\small $^{4}$University of Liverpool, UK} 
{\small $^{*}$Co-First authors}}

\maketitle
\begin{abstract}

In surgical oncology, screening colonoscopy plays a pivotal role in providing diagnostic assistance, such as biopsy, and facilitating surgical navigation, particularly in polyp detection. Computer-assisted endoscopic surgery has recently gained attention and amalgamated various 3D computer vision techniques, including camera localization, depth estimation, surface reconstruction, etc. Neural Radiance Fields (NeRFs) and Neural Implicit Surfaces (NeuS) have emerged as promising methodologies for deriving accurate 3D surface models from sets of registered images, addressing the limitations of existing colon reconstruction approaches stemming from constrained camera movement.

However, the inadequate tissue texture representation and confused scale problem in monocular colonoscopic image reconstruction still impede the progress of the final rendering results. In this paper, we introduce a novel method for colon section reconstruction by leveraging NeuS applied to endoscopic images, supplemented by a single frame of depth map. Notably, we pioneered the exploration of utilizing only one frame depth map in photorealistic reconstruction and neural rendering applications while this single depth map can be easily obtainable from other monocular depth estimation networks with an object scale. Through rigorous experimentation and validation on phantom imagery, our approach demonstrates exceptional accuracy in completely rendering colon sections, even capturing unseen portions of the surface. This breakthrough opens avenues for achieving stable and consistently scaled reconstructions, promising enhanced quality in cancer screening procedures and treatment interventions.
\end{abstract}

\section{Introduction}
\label{sec:intro}

Colorectal cancer stands as the third most frequently diagnosed cancer and the second leading cause of cancer-related mortality \cite{sung2021global}. Timely detection is paramount for favourable prognoses. While various techniques such as virtual colonoscopy exist, optical colonoscopy remains the gold standard for screening and lesion removal. As computer-assisted systems gain prominence in routine endoscopic procedures and present a fertile ground for innovation, typical computer vision approaches are applied including but not limited to depth estimation, 3D reconstruction, and camera localization, benefiting advancements in polyp detection, stenosis assessment, post-intervention diagnosis, and exploration thoroughness evaluation. Nevertheless, achieving precise 3D reconstructions of large colon sections in endoscopic videos remains challenging due to constrained camera movement and the dearth of texture information within the colon.

\begin{figure*}[h]
    \centering
    \includegraphics[width=0.85
    \linewidth]{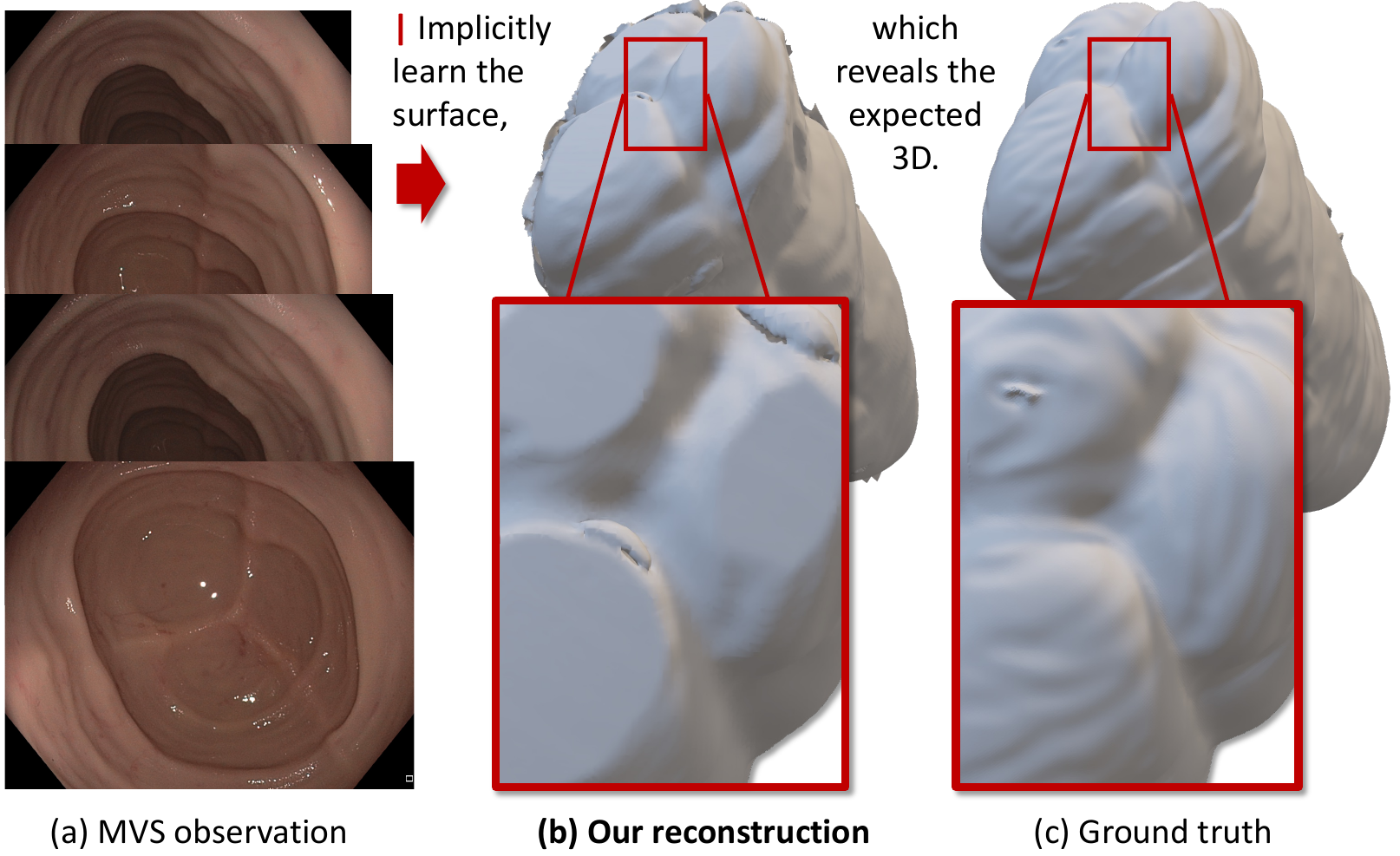}
    \caption{Implicitly reconstructing 3D structures (b) from MVS (multi-view stereo) images (a), which precisely matches the ground truth 3D structures in (c).}
    \label{fig:teaser}
\end{figure*}

To address this challenge, prior research has demonstrated the feasibility of estimating the colon's 3D shape from single images captured during colonoscopies \cite{batlle2022photometric}. Yet, achieving dense reconstructions for large sections necessitates the utilization of multiple images. Since most endoscopes are equipped with a monocular camera, leveraging video sequences using structure-from-motion algorithms emerges as a natural approach \cite{leonard2016image}. Despite advances in image registration techniques \cite{schonberger2016structure} and SLAM \cite{gomez2021sd}, sparse reconstructions remain prevalent, leaving the transition to dense reconstructions unresolved. 

The emergence of Neural Radiance Field (NeRF) \cite{mildenhall2021nerf} networks presents a promising avenue for acquiring implicit 3D representations from image sets. NeRF leverages neural implicit fields for continuous scene representations, demonstrating remarkable success in high-quality view synthesis and 3D reconstruction \cite{martin2021nerf}. 
Instant NGP \cite{muller2022instant} reduced computational costs by introducing versatile new input encoding techniques. 
Recent developments like Gaussian NeRF \cite{kerbl20233d} offer representations using 3D Gaussians, preserving desirable properties of continuous volumetric radiance fields for scene optimization. However, traditional NeRF-based approaches are ill-suited for endoscopic surgery videos. Unlike conventional scenes where cameras capture images from various viewpoints, endoscopic cameras operate within confined spaces, like the cylindrical tunnel of the colon, severely limiting viewing directions and camera movement. Consequently, existing methods, including Neural Implicit Surfaces (NeuS) \cite{wang2021neus}, which represent surfaces as zero-level sets of signed distance functions (SDFs), fail to provide consistent depth mapping in endoscopic scenarios. Therefore, there is a critical need for novel approaches tailored specifically to the unique challenges posed by endoscopic environments.

Our paper presents a new approach to incorporating depth information into NeuS for endoscopy reconstruction. Instead of relying on a dedicated dataset for training a depth estimator \cite{huang2021self, huang2022self, huang2022simultaneous} tailored to surgical scenes, we leverage an existing depth estimator trained on nature scenes. By directly inferring depth maps from this pre-trained estimator and rescaling them to the endoscopy scenes, we effectively supervise the NeuS model. The intensive experiments show that our method achieves state-of-the-art results and produces accurate depth maps on 2D synthetic rendering and 3D reconstruction of endoscopic surgical scenes. 

\section{Methodology}

Given a set of calibrated multi-view images that capture an object or a static scene with their corresponding camera poses, a set of the structural surface and the appearance $\{\mathbf{S}, \mathbf{C} \}$ of the targeted environment could be learned through the radiance supervision~\cite{wang2021neus, long2022neuraludf, wang2022hf}. The learned set $\{ \mathbf{S}, \mathbf{C} \}$ is represented by the signed distance field (SDF) $f(x): \mathbb{R}^3 \rightarrow \mathbb{R}$ where the value of each element is determined by a 3D position $x$, and a radiance field $c(x, v): \mathbb{R}^3 \times \mathbb{S}^2 \rightarrow \mathbb{R}^3$ which is determined by both the position $x$ and the viewing direction $v \in \mathbb{S}^2$. Aiming at learning more precise zero-crossing surfaces in SDF by jointly training the SDF and the radiance field, we introduce two proposed a factor $\lambda_{\rm r}$ to make the SDF regularization more adaptive to improve the rendering quality and reduce the geometric bias.

\subsection{Neural rendering}

By enforcing the radiance supervision through the 2D image,
NeRF~\cite{mildenhall2021nerf} leverages volume rendering to match the ground truth for every camera pose with the rendered image. 
Specifically, the RGB for every pixel of an image can be generated by sampling $n$ points $\{~r(t_i) = o~+~t_i \cdot v~|~i = 1, \dots, n~\}$ along its camera ray $r$, where $o$ is the center of the camera, $t_i$ is the sampling interval along the ray and $v$ is the view direction. 
By accumulating the radiance field density $\sigma(r(t))$ and the colors $c(r(t), v)$ of the sample points, the color $\mathbf{\hat{C}}$ of the ray could be presented as
\begin{align}
\mathbf{\hat{C}}(r) = \int_{t_\text{n}}^{t_\text{f}} T(t) \cdot \sigma(r(t)) \cdot c(r(t), v) ~ dt 
\label{equ:radiance}
\end{align}
where the transparency $T(t)$ is derived from the volume density $\sigma(r(t))$. $T(t)$ denotes the accumulated transmittance along the ray $r$ from the closest point $t_\text{n}$ to the farthest point $t_\text{f}$ such that
\begin{align}
T(t) = \exp \left( -\int_{t_\text{n}}^{t} \sigma(r(s)) ds \right)~.
\end{align}
Note that $T(t)$ is a monotonic decreasing function with a starting value $T(t_\text{n})$ of 1. 
The product $T(t) \cdot \sigma(r(t))$ is used as a weight $\omega(t)$ in the volume rendering of the radiance in Eq. \eqref{equ:radiance}.

Since the rendering process is differentiable, our model can then
learn the radiance field $c$ from the multi-view images with the loss function that minimizes the color difference between the rendered pixels $\mathbf{\hat{C}}(r)$ with $i \in \{1, \dots, m\}$ and the corresponding ground truth pixels $\mathbf{C}(r)$ without 3D supervision as
\begin{align}
    \mathcal{L}_\text{rgb} = \frac{1}{m} \sum_{i=1}^{m} \left(
    \| \mathbf{\hat{C}}(r) - \mathbf{C}(r) \|_2 + 
    | \mathbf{\hat{C}}(r) - \mathbf{C}(r) |
    \right)
    \label{equ:loss_color}
\end{align} 
%
where $m$ denotes the batch size during training. 
Based on the same input and output, we would further investigate a way to implicitly learn a signed distance field $f$ to extract meshes embedded in Eq. \eqref{equ:radiance} during training.

\subsection{Depth-guided SDF optimization}



Aiming at extracting a 3D mesh from a region of interest in the neural rendering,
it is plausible to get a projection from a signed distance function (SDF) to the radiance field.
Here, we look for a function $\Phi$ that transforms the signed distance function so that it can be used to compute the density-related term $T(t) \sigma(r(t))$ in Eq. \eqref{equ:radiance}. 

We build our solution on top of HF-NeuS~\cite{wang2022hf}, where they set $\Phi(r(t))$ as the transparency $T(t)$.
Notably, the derivative of the transparency function $T(t)$ is the negative weighting function as
\begin{align}
\frac{d(T(t))}{dt} = -T(t) \sigma(r(t))~.
\label{equ:trans_density}
\end{align}
Given such formulation, 
the SDF surface lies on the maximum radiance weight.
The maxima is computed by setting the derivative of the weighting function to zero; thus, 
\begin{align}
\frac{d(T(t) \sigma(r(t)))}{dt} = -\frac{d^2(T(t))}{dt^2} = -\frac{d(T'(t))}{dt} = 0~.
\label{equ:unbias_weight}
\end{align}
%
%
To fulfill the criteria from Eq. \eqref{equ:trans_density} and \eqref{equ:unbias_weight}, HF-NeuS~\cite{wang2022hf} defined the transparency function $T_\text{s}(t)$ as the normalized sigmoid function $\left[1+\exp\left({s \cdot f(r(t))}\right)\right]^{-1}$
with a parameter $s$. The scalar $s$ reveals how strong the SDF is related to the radiance field, which is usually increasing during training. In the initial stage of training, a small $s$ relaxes the connection between SDF and radiance such that the radiance parametric model is optimized without a dependency on the plausible SDF, where the view-conditioned depth $\mathbf{Z}(r_i)$ for each ray $r_i$ matches the ground truth depth $\mathbf{\hat{Z}}(r_i)$. 

To parameterize SDF, a signed distance function $f(x)$ is differentiable almost everywhere, while its gradient $\grad f(x)$ satisfies the Eikonal equation $\|\grad f(x)\|_2 = 1$.
This implies that an SDF can be trained with the Eikonal regularization. 
According to IGR~\cite{gropp2020implicit}, we enforce an Eikonal loss as a regularizer to make an implicit field act as a signed distance field. However, we discovered that training with a fixed Eikonal regularization~\cite{wang2021neus, wang2022hf, zhao2022human} can lead to a sub-optimal convergence where the SDF has converged based on the Eikonal regularization but the rendering can still be further optimized. 
Such a problem harms the improvement of the RGB rendering because the weights are projected from a wrong SDF value, which makes the extracted mesh miss the detailed structures.

\begin{table*}[h!]
\caption{Quantitative results. The best result in each small sequence is presented in bold.}
\resizebox{\textwidth}{!}{
\begin{tabular}{|c|c|c|c|c|c|c|c|c|c|c|c|c|c|}

\hline
Method &Item &c1\_a &c1\_b  & c2\_a & c2\_b & c2\_c  & c3\_a & c4\_a &c4\_b  &d4\_a &s1\_a &s2\_a \\ 
\hline
Gaussian NeRF \cite{kerbl20233d}
&psnr  &32.84  &28.19 &30.24 &\textbf{31.07} &28.25 &26.53 &28.41 &32.03 &29.26 
 &\textbf{36.27}  &35.22  \\
\hline

Vanilla NeRF \cite{mildenhall2021nerf}
&psnr  &32.71  &30.68 &31.35 &29.32 &\textbf{32.32} &\textbf{33.42} &32.35 &33.70 &28.65 
 &35.92  &31.00  \\

\hline
Ours
 & psnr  & \textbf{34.27} & \textbf{31.02} & \textbf{33.78} & 31.04 & 31.02 & 32.75 & \textbf{33.62} & \textbf{34.19} & \textbf{31.01}
 & 31.24 & \textbf{35.88} \\
 
 \hline
 
 
\hline
\end{tabular}}
\label{quantitive results}
\end{table*}

To solve the problem of the reconstruction failures on low-textured structures, \textit{i.e.} internal surface of organ, we found that the capability of rendering such structures from the radiance field should be ensured even when the SDF does not contain the accorded structures so that the gradient could be back-propagated from the radiance field to the SDF later on.
Given the overall loss function $\mathcal{L}_\text{total} = \mathcal{L}_\text{rgb} + \mathcal{L}_\text{sdf}$~, RaNeuS~\cite{wang2024raneus} proposed to adaptively weight the Eikonal regularization to optimize the SDF as
\begin{align}
    \mathcal{L}_\text{sdf} = \frac{\lambda_\text{E}}{m n} \sum_{i=1}^{m} \lambda_\text{r}(r_i) \sum_{j=1}^{n} \left(\| \text{n}_{ij} \|_2 - 1 \right)^2 ~,
    \label{equ:loss_eikonal}
\end{align}
where a ray-wise weight $\lambda_\text{r}(r_i)$ for the $i$-th ray $r_i$ is set to be 
\begin{align}
    \lambda_\text{r}(r_i) = \frac{\alpha}{d_\text{r}(r_i) + \alpha} ~.
    \label{equ:lambda_r}
\end{align}
We propose to present the ray-wise weight $\lambda_\text{r}(r_i)$ considering a depth bias instead. Given the depth value $\mathbf{Z}(r_i)$ determined by the zero-crossing point sampled from ray $r_i$, $d_\text{r}(r_i)$ is the depth distance $\| \mathbf{\hat{Z}}(r_i) - \mathbf{Z}(r_i) \|_2$ regarding ray $r_i$, and $\alpha$ is a positive hyperparameter that is set to be smaller than 1, \textit{e.g.} $1 \cdot 10^{-3}$. 
Notice that a precise per-ray depth supervision $\mathbf{\hat{Z}}(r_i)$ is not necessarily supplemented, while the depth distance $d_\text{r}(r_i)$ is not optimized directly here. 
Given a general depth estimation model such as a DPT Hybrid model~\cite{liu2024dpdformer}, we rescale the predicted depth according to a reconstructed MVS point cloud using COLMAP~\cite{schoenberger2016sfm, schoenberger2016mvs} to represent an approximation of $\mathbf{\hat{Z}}(r_i)$. 
Practically, the gradients back-propagated to the zero crossing point $\mathbf{Z}(r_i)$ are disabled during training. Consequently, the Eikonal regularization will be relaxed when the geometric precision determined by the metric of $d_\text{r}(r_i)$ is unsatisfying.

We follow the remaining hyper-parameter setups of RaNeuS~\cite{wang2024raneus}, where the approximated normal $\text{n} = \grad f(r(\cdot))$ is the derivative of $f(r(\cdot))$, and $\lambda_\text{E}$ is typically set to be 0.1 as mentioned by IGR~\cite{gropp2020implicit} and NeuS~\cite{wang2021neus}. 

To avoid extreme values for different scenes, $d_\text{r}(r)$ is set to be zero when a ray does not hit the foreground SDF space. 
A converged model is respected to end up with the typical Eikonal regularization when there is precise surface geometry measured by a small $d_\text{r}(r)$. Thus, in contrast to \cite{gropp2020implicit}, our approach is more adaptive to the changes in both the rendering and SDF optimizations; thus, relaxing the restrictions from the predefined parameters.

\section{Experimental Results}

Our method was trained and assessed using C3VD \cite{bobrow2023colonoscopy}, a dataset comprising small video sequences obtained from real wide-angle colonoscopies at a resolution of 1350 × 1080. These sequences traverse four distinct colon phantoms, including the colon cecum, descending, sigmoid, and transcending regions. Video lengths vary from 61 to 1142 frames, with per-frame camera poses utilized. We divided each scene into training, testing, and validation sets in a 6:2:2 ratio. Additionally, we compared our model's performance with that of Gaussian NeRF \cite{kerbl20233d} and vanilla NeRF. 

The results, presented in Table \ref{quantitive results}, were evaluated based on RGB reconstruction quality, measured using peak signal-to-noise ratio (PSNR). Notably, our method consistently outperforms or matches the performance of Gaussian NeRF \cite{kerbl20233d} and vanilla NeRF \cite{mildenhall2021nerf} across most scenarios, as indicated in Table \ref{quantitive results}.

\section{Conclusion}
In conclusion, we have introduced an approach to enhance the reconstruction and rendering quality of endoscopic images by integrating single-frame depth-guided SDF optimization. We employed adaptive regularization to mitigate geometric bias, and our method's efficacy was demonstrated through PSNR metrics. Remarkably, even without depth guidance, our method surpassed the performance of Gaussian NeRF \cite{kerbl20233d} and vanilla NeRF \cite{mildenhall2021nerf}. Moreover, the incorporation of depth guidance resulted in further improvements in the quality of neural rendering.

\section{Acknowledgements}
This work is supported by EU-H2020 EndoMapper GA863146.

{
    \small
    \bibliographystyle{ieeenat_fullname}
    \bibliography{main}

\begin{thebibliography}{22}
\providecommand{\natexlab}[1]{#1}
\providecommand{\url}[1]{\texttt{#1}}
\expandafter\ifx\csname urlstyle\endcsname\relax
  \providecommand{\doi}[1]{doi: #1}\else
  \providecommand{\doi}{doi: \begingroup \urlstyle{rm}\Url}\fi

\bibitem[Batlle et~al.(2022)Batlle, Montiel, and Tard{\'o}s]{batlle2022photometric}
V{\'\i}ctor~M Batlle, Jos{\'e}~MM Montiel, and Juan~D Tard{\'o}s.
\newblock Photometric single-view dense 3d reconstruction in endoscopy.
\newblock In \emph{IROS}, 2022.

\bibitem[Bobrow et~al.(2023)Bobrow, Golhar, Vijayan, Akshintala, Garcia, and Durr]{bobrow2023colonoscopy}
Taylor~L Bobrow, Mayank Golhar, Rohan Vijayan, Venkata~S Akshintala, Juan~R Garcia, and Nicholas~J Durr.
\newblock Colonoscopy 3d video dataset with paired depth from 2d-3d registration.
\newblock \emph{Medical image analysis}, 2023.

\bibitem[G{\'o}mez-Rodr{\'\i}guez et~al.(2021)G{\'o}mez-Rodr{\'\i}guez, Lamarca, Morlana, Tard{\'o}s, and Montiel]{gomez2021sd}
Juan~J G{\'o}mez-Rodr{\'\i}guez, Jos{\'e} Lamarca, Javier Morlana, Juan~D Tard{\'o}s, and Jos{\'e}~MM Montiel.
\newblock Sd-defslam: Semi-direct monocular slam for deformable and intracorporeal scenes.
\newblock In \emph{ICRA}, 2021.

\bibitem[Gropp et~al.(2020)Gropp, Yariv, Haim, Atzmon, and Lipman]{gropp2020implicit}
Amos Gropp, Lior Yariv, Niv Haim, Matan Atzmon, and Yaron Lipman.
\newblock Implicit geometric regularization for learning shapes.
\newblock \emph{arXiv preprint arXiv:2002.10099}, 2020.

\bibitem[Huang et~al.(2021)Huang, Zheng, Nguyen, Tuch, Vyas, Giannarou, and Elson]{huang2021self}
Baoru Huang, Jian-Qing Zheng, Anh Nguyen, David Tuch, Kunal Vyas, Stamatia Giannarou, and Daniel~S Elson.
\newblock Self-supervised generative adversarial network for depth estimation in laparoscopic images.
\newblock In \emph{MICCAI}, 2021.

\bibitem[Huang et~al.(2022{\natexlab{a}})Huang, Nguyen, Wang, Wang, Mayer, Tuch, Vyas, Giannarou, and Elson]{huang2022simultaneous}
Baoru Huang, Anh Nguyen, Siyao Wang, Ziyang Wang, Erik Mayer, David Tuch, Kunal Vyas, Stamatia Giannarou, and Daniel~S Elson.
\newblock Simultaneous depth estimation and surgical tool segmentation in laparoscopic images.
\newblock \emph{TMRB}, 2022{\natexlab{a}}.

\bibitem[Huang et~al.(2022{\natexlab{b}})Huang, Zheng, Nguyen, Xu, Gkouzionis, Vyas, Tuch, Giannarou, and Elson]{huang2022self}
Baoru Huang, Jian-Qing Zheng, Anh Nguyen, Chi Xu, Ioannis Gkouzionis, Kunal Vyas, David Tuch, Stamatia Giannarou, and Daniel~S Elson.
\newblock Self-supervised depth estimation in laparoscopic image using 3d geometric consistency.
\newblock In \emph{MICCAI}, 2022{\natexlab{b}}.

\bibitem[Kerbl et~al.(2023)Kerbl, Kopanas, Leimk{\"u}hler, and Drettakis]{kerbl20233d}
Bernhard Kerbl, Georgios Kopanas, Thomas Leimk{\"u}hler, and George Drettakis.
\newblock 3d gaussian splatting for real-time radiance field rendering.
\newblock \emph{ACM Transactions on Graphics}, 2023.

\bibitem[Leonard et~al.(2016)Leonard, Reiter, Sinha, Ishii, Taylor, and Hager]{leonard2016image}
Simon Leonard, Austin Reiter, Ayushi Sinha, Masaru Ishii, Russell~H Taylor, and Gregory~D Hager.
\newblock Image-based navigation for functional endoscopic sinus surgery using structure from motion.
\newblock In \emph{Medical Imaging}, 2016.

\bibitem[Liu et~al.(2024)Liu, Yang, Zuo, and Zang]{liu2024dpdformer}
Chunpu Liu, Guanglei Yang, Wangmeng Zuo, and Tianyi Zang.
\newblock Dpdformer: a coarse-to-fine model for monocular depth estimation.
\newblock \emph{ACM Transactions on Multimedia Computing, Communications and Applications}, 2024.

\bibitem[Long et~al.(2022)Long, Lin, Liu, Liu, Wang, Theobalt, Komura, and Wang]{long2022neuraludf}
Xiaoxiao Long, Cheng Lin, Lingjie Liu, Yuan Liu, Peng Wang, Christian Theobalt, Taku Komura, and Wenping Wang.
\newblock Neuraludf: Learning unsigned distance fields for multi-view reconstruction of surfaces with arbitrary topologies.
\newblock \emph{arXiv preprint arXiv:2211.14173}, 2022.

\bibitem[Martin-Brualla et~al.(2021)Martin-Brualla, Radwan, Sajjadi, Barron, Dosovitskiy, and Duckworth]{martin2021nerf}
Ricardo Martin-Brualla, Noha Radwan, Mehdi~SM Sajjadi, Jonathan~T Barron, Alexey Dosovitskiy, and Daniel Duckworth.
\newblock Nerf in the wild: Neural radiance fields for unconstrained photo collections.
\newblock In \emph{CVPR}, 2021.

\bibitem[Mildenhall et~al.(2021)Mildenhall, Srinivasan, Tancik, Barron, Ramamoorthi, and Ng]{mildenhall2021nerf}
Ben Mildenhall, Pratul~P Srinivasan, Matthew Tancik, Jonathan~T Barron, Ravi Ramamoorthi, and Ren Ng.
\newblock Nerf: Representing scenes as neural radiance fields for view synthesis.
\newblock \emph{Communications of the ACM}, 2021.

\bibitem[M{\"u}ller et~al.(2022)M{\"u}ller, Evans, Schied, and Keller]{muller2022instant}
Thomas M{\"u}ller, Alex Evans, Christoph Schied, and Alexander Keller.
\newblock Instant neural graphics primitives with a multiresolution hash encoding.
\newblock \emph{ACM Transactions on Graphics (ToG)}, 2022.

\bibitem[Sch\"{o}nberger and Frahm(2016)]{schoenberger2016sfm}
Johannes~Lutz Sch\"{o}nberger and Jan-Michael Frahm.
\newblock Structure-from-motion revisited.
\newblock In \emph{CVPR}, 2016.

\bibitem[Schonberger and Frahm(2016)]{schonberger2016structure}
Johannes~L Schonberger and Jan-Michael Frahm.
\newblock Structure-from-motion revisited.
\newblock In \emph{CVPR}, 2016.

\bibitem[Sch\"{o}nberger et~al.(2016)Sch\"{o}nberger, Zheng, Pollefeys, and Frahm]{schoenberger2016mvs}
Johannes~Lutz Sch\"{o}nberger, Enliang Zheng, Marc Pollefeys, and Jan-Michael Frahm.
\newblock Pixelwise view selection for unstructured multi-view stereo.
\newblock In \emph{ECCV}, 2016.

\bibitem[Sung et~al.(2021)Sung, Ferlay, Siegel, Laversanne, Soerjomataram, Jemal, and Bray]{sung2021global}
Hyuna Sung, Jacques Ferlay, Rebecca~L Siegel, Mathieu Laversanne, Isabelle Soerjomataram, Ahmedin Jemal, and Freddie Bray.
\newblock Global cancer statistics 2020: Globocan estimates of incidence and mortality worldwide for 36 cancers in 185 countries.
\newblock \emph{CA: a cancer journal for clinicians}, 2021.

\bibitem[Wang et~al.(2021)Wang, Liu, Liu, Theobalt, Komura, and Wang]{wang2021neus}
Peng Wang, Lingjie Liu, Yuan Liu, Christian Theobalt, Taku Komura, and Wenping Wang.
\newblock Neus: Learning neural implicit surfaces by volume rendering for multi-view reconstruction.
\newblock \emph{arXiv preprint arXiv:2106.10689}, 2021.

\bibitem[Wang et~al.(2022)Wang, Skorokhodov, and Wonka]{wang2022hf}
Yiqun Wang, Ivan Skorokhodov, and Peter Wonka.
\newblock Hf-neus: Improved surface reconstruction using high-frequency details.
\newblock \emph{NeurIPS}, 2022.

\bibitem[Wang et~al.(2024)Wang, Tan, Navab, and Tombari]{wang2024raneus}
Yida Wang, David~Joseph Tan, Nassir Navab, and Federico Tombari.
\newblock Raneus: Ray-adaptive neural surface reconstruction.
\newblock In \emph{2024 International Conference on 3D Vision (3DV)}. IEEE, 2024.

\bibitem[Zhao et~al.(2022)Zhao, Jiang, Yao, Zhang, Wang, Dai, Zhong, Zhang, Wu, Xu, et~al.]{zhao2022human}
Fuqiang Zhao, Yuheng Jiang, Kaixin Yao, Jiakai Zhang, Liao Wang, Haizhao Dai, Yuhui Zhong, Yingliang Zhang, Minye Wu, Lan Xu, et~al.
\newblock Human performance modeling and rendering via neural animated mesh.
\newblock \emph{ACM Transactions on Graphics (TOG)}, 2022.

\end{thebibliography}
}

\end{document}